\documentclass[letterpaper, 10 pt, conference]{ieeeconf}  

\IEEEoverridecommandlockouts                              

\usepackage{amsmath}
\usepackage{graphicx}
\usepackage{amssymb}
\usepackage{subcaption}
\usepackage{booktabs} 
\usepackage{gensymb}
\usepackage{color,soul}
\usepackage{adjustbox}
\usepackage{tikz}
\usepackage{cleveref}
\usepackage{multirow}
\usepackage{url}
\usepackage{colortbl}
\title{\LARGE \bf
VE2VF: Vision-Enabled to Vision-Free Distillation via Real-world Reinforcement Learning for Robust Contact-Rich Manipulation
}

\author{Victor Kowalski$^{1}$, Chengxi Li$^{1}$, and Dongheui Lee$^{1,2}$
\thanks{This work was supported by the European Union project INVERSE under
grant agreement No. 101136067.}
\thanks{$^{1}$Victor Kowalski, Chengxi Li, and Dongheui Lee are with
Autonomous Systems, Technische Universität Wien (TU Wien), Vienna, Austria 
        {\tt\small \{victor.martins, chengxi.li, dongheui.lee\}@tuwien.ac.at}}  %
\thanks{$^{2}$Dongheui Lee is also with the Institute of Robotics and Mechatronics
(DLR), German Aerospace Center, Wessling, Germany.}%
}

\begin{document}

\maketitle
\thispagestyle{empty}
\pagestyle{empty}

\begin{abstract}
When using reinforcement learning (RL) for contact-rich robotic manipulation, vision can provide task-relevant information that complements robot proprioception. However, vision-enabled policies tend to overfit to the visual conditions seen during training, limiting their robustness and transferability.
We present a human-in-the-loop RL framework that employs teacher-student distillation to achieve robust performance across multiple task variants, trained entirely in the real world without requiring domain randomization or data augmentation. A vision-enabled teacher distills its knowledge into a vision-free student that relies solely on pose, twist, and wrench sensing, combining fast training with strong task generalization. On the real-world NIST assembly benchmark board, our approach achieves 95\% overall success after approximately 50 minutes of training on 3 representative tasks, including robust generalization to 8 unseen task variants. Fine-tuning with distillation achieves full success on the most challenging task. We demonstrate that the resulting policies outperform baselines in both robustness and adaptability. Page: \url{ https://tuwien-asl.github.io/VE2VF/}.
\end{abstract}

\section{Introduction}   
Robotic assembly is a longstanding challenge involving high-precision contact-rich interactions. In addition, high-mix, low-volume assembly settings require adaptability to diverse parts and environments. Traditional model-based approaches often fail in this domain since accurately modeling contact dynamics is notoriously difficult, and even small errors can lead to unstable behavior~\cite{contactrichsurvey}. To address these limitations, learning-based approaches have been increasingly explored. Different paradigms of robotic manipulation learning offer distinct advantages and drawbacks. Imitation learning (IL), for instance, provides a data-efficient way to bootstrap policies from demonstrations~\cite{diffusion,Zhao-RSS-23}, achieving strong results in structured tasks. However, IL policies are typically limited to demonstration distribution and struggle in novel situations~\cite{pmlr-v9-ross10a}. This is particularly problematic in contact-rich manipulation, where small deviations can cascade into failure. Reinforcement learning (RL), on the other hand, enables policy improvement through trial and error \cite{deeprlsurvey}. Physics-based simulators~\cite{mittal2025isaaclab,mujoco} facilitate robot data generation, but the gap between simulation and reality is particularly pronounced for contact-rich interactions, where accurate modeling of geometric and dynamic properties is challenging~\cite{forge}. Human-in-the-loop RL (HIL-RL)~\cite{hilserl} is a sample-efficient alternative that leverages human demonstrations and interventions to guide exploration, making real-world training practical. 

\begin{figure}[t]
  \centering
  \includegraphics[width=0.9\linewidth]{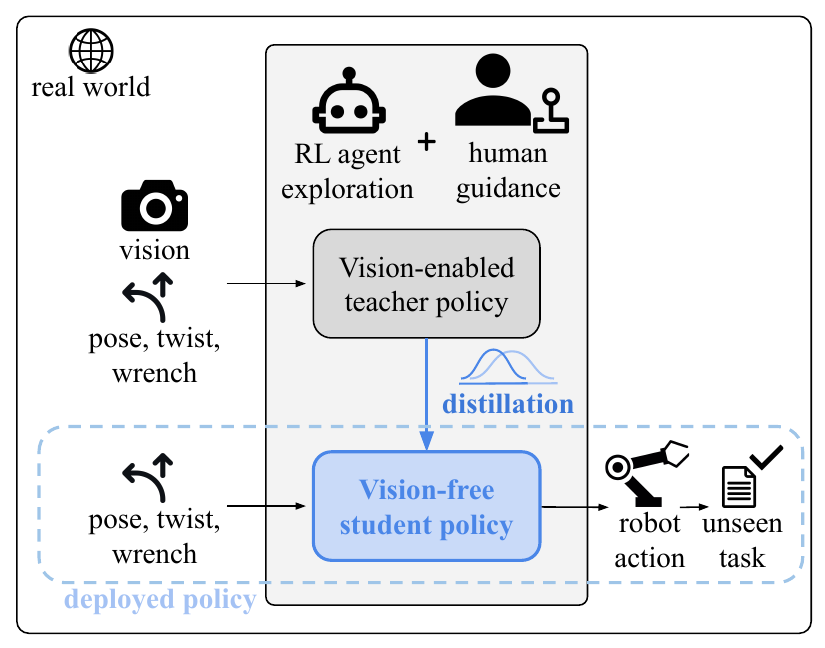}
 \caption{An overview of our proposed approach \textit{VE2VF}. A vision-enabled teacher policy is trained via human-in-the-loop RL on a representative set of tasks, then distilled into a vision-free student policy that does not rely on the environment's appearance, enabling robust performance and generalization to unseen tasks.}
  \vspace{-0.4cm}
  \label{overview fig}
\end{figure}

Visual input has been a core component of policy learning~\cite{levine2016end}. However, vision-enabled policies often rely on arbitrary image features, which limit their generalization across varying visual conditions. Changes in lighting, object textures, or backgrounds can significantly impact performance~\cite{stone2021distracting}. We therefore propose a vision-free HIL-RL framework that maps sensed poses, velocities, and force-torques (wrenches) to actions, remaining invariant to visual shifts. Yet learning such policies directly from scratch is challenging due to partial observability and the difficulty of exploration without visual guidance. To overcome this, our \textit{VE2VF} framework follows a two-stage approach, as shown in Fig.~\ref{overview fig}. First, a vision-enabled teacher policy is trained to benefit from rich perceptual feedback. Then, we employ knowledge distillation to transfer the acquired skills to a vision-free student policy. While teacher-student distillation has been widely adopted in robot learning~\cite{lee2020learning, miki2022learning,azulay2025visuotactile,transic,viral}, those works typically focus on sim-to-real transfer. Our motivation differs: for contact-rich manipulation, visual input can serve as a distractor from the essential forces and geometric relationships that govern task success, so distillation here helps shed an unreliable modality while retaining the contextual benefits that vision provides during training. In an optional third stage, policy fine-tuning can be performed to adapt to challenging, unseen tasks. 

In summary, our work makes the following contributions:
\begin{itemize}
\item \textbf{Cross-modal distillation framework}: We present a method that distills vision-enabled teacher policies into vision-free student policies, retaining the contextual information of visual learning while producing proprioceptive policies invariant to visual conditions.
\item \textbf{Efficient skill generalization}: We demonstrate that co-training on a small set of representative tasks enables robust transfer to unseen insertion scenarios, with our policies achieving zero-shot success or complete success after fine-tuning with distillation.
\item \textbf{Real-world validation}: Through comprehensive experiments on contact-rich insertion tasks, we show that our approach outperforms baseline methods in both robustness and adaptability, while completing training in approximately 50 minutes of robot interaction.
\end{itemize}

\section{Related Work}
This work leverages real-world RL with human guidance to acquire contact-rich robotic skills in a sample-efficient manner. Furthermore, we achieve greater policy robustness by removing the visual observation modality during deployment and relying solely on proprioceptive information. Accordingly, we review related work in reinforcement learning for contact-rich manipulation and multimodal sensory observations for policy robustness.

\subsection{Reinforcement Learning for Contact-rich Manipulation}

Several approaches have addressed contact-rich manipulation by training RL policies in physics-based simulators and deploying them in the real world \cite{Tang-RSS-23}, \cite{tang2024automate}, \cite{Guo-ICLR-25-SRSA}, \cite{azulay2025visuotactile}, \cite{transic}. While these methods enable easy data generation and achieve significant performance, they require extensive simulation infrastructure and remain sensitive to sim-to-real gaps. Recent works on real-world RL propose learning contact-rich manipulation directly in the real world but require many environment transitions \cite{Nguyen-2024-SymmetrySoftWrist} or task constraints to enable efficient exploration \cite{towardsDLR}.  Imitation learning approaches such as DMPs \cite{DMP} or the more recent Diffusion Policy \cite{diffusion} evade the exploration issue by mimicking expert demonstrations. The REASSEMBLE dataset \cite{sliwowski2025reassemble} provides expert demonstrations of contact-intensive assembly tasks but remains hardware- and environment-specific, limiting its applicability in other deployment scenarios. Inevitably, IL suffers from compounding errors outside the demonstration distribution \cite{pmlr-v9-ross10a}, which has inspired the augmentation of demonstration-derived controllers with learned residual corrections, thereby improving precision and enabling recovery behaviors. rLfD \cite{residualfromdemo} refines a DMP policy using an RL-based residual, while ResiP \cite{Ankile-2024-ResiP} uses a base Diffusion Policy instead of DMPs. Alternatively, human-in-the-loop learning addresses this issue through online expert intervention, allowing the robot to recover from undesirable states encountered during policy execution. Human-gated DAgger \cite{kelly2019hg} is a behavior cloning method in which a human supervisor determines when to correct the robot’s actions, and the policy is retrained with the added data. HIL-SERL \cite{hilserl} introduces this idea in the real-world RL context and shows efficient acquisition of contact-rich manipulation skills directly on hardware. 

\subsection{Multimodal sensory observations for policy robustness}

Leveraging diverse observation modalities can lead to more robust policy learning \cite{midlevel,msdp,seehearfeel,makingsense, vtt,factr,patil2025factorizing,azulay2025visuotactile}. Properly chosen mid-level visual representations as observation inputs have been shown to improve policy robustness \cite{midlevel}. Later works have added other modalities, such as haptic sensing, and trained encoders to combine different observation inputs into a latent vector and optimize downstream task performance \cite{msdp,seehearfeel,makingsense,vtt}. Instead of focusing on representation learning, other works train policies by strictly de-prioritizing vision, relying more on haptic information \cite{factr,patil2025factorizing}. Similarly, we want to avoid relying on vision during deployment and instead employ a distillation technique common in the sim-to-real domain ~\cite{lee2020learning, miki2022learning,transic,viral,azulay2025visuotactile}, in which a teacher policy using privileged observation modalities from the simulation serves as an expert to be mimicked by a student using observations available in the real world. In our case, both the teacher and the student are trained in the real world, and we treat vision as the privileged modality since it provides useful task context, but, as our experiments highlight, it is unreliable during deployment.

\section{Preliminary} \label{sec preliminary}



\subsection{Problem Formulation} \label{sec problem formulation}
Each robotic contact-rich task can be formulated as a Markov Decision Process (MDP) $\mathcal{M}=(\mathcal{S}, \mathcal{A}, \rho, \mathcal{P}, r, \gamma)$, where $s\in S$ is the state observation, $a\in\mathcal{A}$ is the action, $\rho\left(s_0\right)$ is a distribution over robots' initial states, $\mathcal{P}$ is the unknown and potentially stochastic transition probabilities that depend on the system dynamics, and $r : S\times\mathcal{A} \rightarrow \mathbb{R}$ is the reward function that encodes the task, and $\gamma$ is a discount factor $0 < \gamma < 1$ used to prioritize earlier rewards. An optimal policy $\pi$ is one that maximizes the cumulative expected value of the reward, i.e.,  $ E=[\sum_{t=0}^H \gamma^tr(s_t,a_t)]$, where the expectation is taken with respect to the initial state distribution, transition probabilities, and policy $\pi$.

\subsection{Policy optimization} \label{sec sac opt}

Our policy is learned via off-policy max-entropy RL \cite{SAC}, which provides the sample efficiency necessary for real-world training. A critic model $Q_\phi(s,a)$ is trained to estimate the value of state-action pairs across the search space, while an actor (policy) model $\pi_\theta(a|s)$ stochastically chooses the action to take at a given state, raising the probability of actions deemed as good by the critic. A dynamic temperature parameter $\alpha$ ensures that the actor does not prematurely converge to deterministic, locally optimal actions, thereby incentivizing exploration. The critic model parameters are updated to minimize the temporal difference error over past transitions stored in the buffer $\mathcal{D}$:
\begin{equation}
\mathcal{L}_Q(\phi) = \mathbb{E}_{(s,a,r,s') \sim \mathcal{D}} \left[ \left( Q_\phi(s,a) - y \right)^2 \right]
\label{critic loss}
\end{equation}
where:
\begin{equation}
y = r + \gamma \left(Q_{{\phi}}(s', a') - \alpha \log \pi_\theta(a'|s') \right),
\end{equation}
with $a' \sim \pi_\theta(\cdot|s')$ sampled from the current policy. Therefore, the critic model is updated to reflect immediate rewards and the time-discounted value of the future state, minus a measure of the policy's certainty in taking the sampled action in that future state. The actor's parameters are updated to maximize the probability of taking high-value actions, again regularized by a measure of certainty in taking the action, ensuring exploration. That is achieved by minimizing:
\begin{equation}
\resizebox{0.96\width}{!}{$\displaystyle
    \mathcal{L}_\pi(\theta) = - \mathbb{E}_{s \sim \mathcal{D}} \left[ \mathbb{E}_{a \sim \pi_\theta(\cdot|s)} \left[ Q_\phi(s,a) - \alpha \log \pi_\theta(a|s)\right] \right]
$.}
\label{actor_loss}
\end{equation}

\subsection{Human-in-the-Loop Reinforcement Learning} \label{sec rlpd}

We train our policy in a human-in-the-loop setting, following the approach introduced in HIL-SERL \cite{hilserl}. The RLPD (Reinforcement Learning with Prior Data) algorithm~\cite{ball2023efficient} is used to symmetrically sample transitions from policy-generated data $\mathcal{D}_{\text{P}}$ and human-generated data $\mathcal{D}_{\text{H}}$, which we represent as $(\cdot)\sim \mathcal{D}_{\text{HIL}}$. The human-generated data comes in two forms: first, offline human demonstrations are collected; second, a human can teleoperate the robot during online interaction, overwriting the actions produced by the RL agent. In this setting, the algorithm learns from both autonomous policy rollouts and human demonstrations and interventions, enabling efficient learning from human guidance while maintaining continuous improvement through self-generated experience.

\section{Methodology} \label{methodology}

\begin{figure*}[t]
  \centering
  \includegraphics[trim=0pt 4pt 0pt 4pt, clip, width=0.61\linewidth]{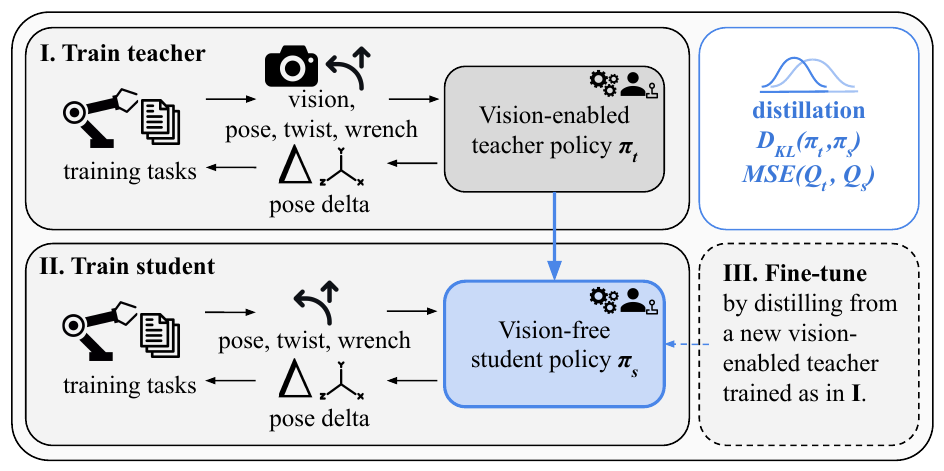}
  \caption{Stages of our \textit{VE2VF} method: in \textbf{I}, a vision-enabled teacher is trained using all input modalities to establish an expert prior policy; in \textbf{II}, the vision-enabled teacher is distilled into a vision-free student that relies only on proprioceptive observations, enabling broader applicability. Still, some novel tasks may present characteristics not sufficiently represented in the training tasks, for which we apply an optional stage \textbf{III}, in which a new vision-enabled teacher is trained as in \textbf{I}, and the previously trained student is fine-tuned with distillation from this new teacher.}
  \vspace{-0.4cm}
  \label{method diagram}
\end{figure*}

We propose a real-world pipeline for learning generalizable policies. Section~\ref{section policy definition} describes our MDP implementation details. The policy learning process starts with the selection of training tasks discussed in Sec. \ref{sec Task curriculum selection}. Using this training task set, a vision-enabled policy is trained using all input modalities, as described in Sec. \ref{sec Vision-enabled teacher training}. To overcome the environment appearance dependency, we distill the vision-enabled policy's knowledge into a vision-free policy that takes only proprioceptive inputs, as explained in Sec. \ref{sec vision-free student distillation}. Finally, Sec.~\ref{sec zeroshot fewshot} shows how the vision-free policy can be used for novel tasks in a zero-shot manner or with fine-tuning. Figure~\ref{method diagram} depicts the components of our method \textit{VE2VF}. 


\subsection{MDP definition}\label{section policy definition}

In this section, the state observation space, the action space, the reward function, and the initial state distribution of our policies are detailed:

\subsubsection{State Observation space $\mathcal{S}$} The state observations consist of 
    $o_t = \big( \mathcal{I}_t, \; x_t \big)$, where
    \begin{equation}
        \mathcal{I}_t = \{ I_t^{(k)} \mid k = 1, \dots, n \}, 
    \quad I_t^{(k)} \in \mathbb{R}^{H \times W \times 3},
    \end{equation}
    is a set of images from $n$ cameras,
    \begin{equation}
    x_t^{\text{prop}} = \big( p_t, \; v_t, \; f_t, \; \tau_t \big),
    \end{equation}is the proprioceptive state vector composed of the end-effector pose $p_t \in \mathbb{R}^6$, velocity $v_t \in \mathbb{R}^6$, force $f_t \in \mathbb{R}^3$, and torque $\tau_t \in \mathbb{R}^3$, represented with respect to the end-effector frame. 
         
    \subsubsection{Action Space $\mathcal{A}$} 
    At each time step $t$, the robot action $\mathbf{a}_t$ is defined as a pose displacement $\Delta \mathbf{p}_t$ expressed in the local coordinate frame of the end-effector. The commands are executed on the manipulator through a low-level impedance controller, which ensures compliant realization of the desired motion. The displacement $\Delta \mathbf{p}_t$ is clipped to keep the robot within a bounded region in the environment.

    \subsubsection{Reward Space $\mathcal{R}$}
    A key aspect of RL systems is the reward function $r_t$, which guides the learning process and evaluates policy performance. Previous work has often employed reward shaping $r^{\text{shaping}}_t$ to accelerate learning; however, this approach is typically task-specific and time-consuming to design. In multi-task settings, it is infeasible to carefully design reward shaping for each task, and such rewards cannot be easily transferred to new tasks.  

    To address this, we use a binary reward classifier $C(s_t)$, where $s_t$ denotes the state at time $t$. The reward is then defined as  
    \[
    r_t =
    \begin{cases}
    1, & \text{if } C(s_t) = 1 \;\; \text{(task success)} \\
    0, & \text{otherwise},
    \end{cases}
    \]
    which corresponds to a sparse reward function that only assigns a positive signal upon task completion. By combining this sparse reward with human demonstrations and human corrections, we obtain a direct and effective scheme for training across multiple tasks.

 \subsubsection{Initial state distribution}
 Each episode is initialized to a random pose within the action space.

\subsection{Training tasks selection} \label{sec Task curriculum selection}

Selecting a representative and diverse set of training tasks is fundamental to ensuring that our vision-free policy generalizes broadly rather than overfitting to a narrow set of contact dynamics. With that in mind, we select a set of contact-rich manipulation tasks with distinct geometrical profiles and interaction stiffness. Our policies learn these tasks simultaneously by looping through them at each $n$-th policy episode.

\subsection{Vision-enabled teacher training} \label{sec Vision-enabled teacher training}

We start by training the vision-enabled teacher policy $\pi^{\text{t}}$ via the human-in-the-loop reinforcement learning (HIL-RL) framework described in Sec.~\ref{sec preliminary}.

We use the full multimodal observations $o^\text{t} = \big(\mathcal{I}, \; x^{\text{t}}, \; z \big)$, 
where the modified proprioceptive state
\begin{equation}
     x^{\text{t}} = \big(  T_{p_0}(p), \; v, \; f, \; \tau \big)
\end{equation}
represents poses relatively to the initial pose of the observed episode via a homogeneous transformation $T_{p_0}(\cdot)$. 

The vision-enabled policy's critic loss reads as:
\begin{equation}
\mathcal{L}_{Q}^\text{t}(\phi) = \mathbb{E}_{(o^{\text{t}},a,r,{o^{\text{t}}}') \sim \mathcal{D}^\text{t}_\text{HIL}} \left[ \left( Q^\text{t}_{\phi}(o^{\text{t}}, a) - y \right)^2 \right],
\label{critic loss}
\end{equation}
where:
\begin{equation}
y = r + \gamma \left( \min_{i=1,2} Q^\text{t}_{\bar{\phi}_i}({o^{\text{t}}}', a') - \alpha \log \pi^\text{t}_\theta(a'|{o^{\text{t}}}') \right).
\end{equation}

The actor loss reads as:
\begin{equation}
\mathcal{L}^\text{t}_\pi(\theta) = \mathbb{E}_{o \sim \mathcal{D}^\text{t}_\text{HIL}} \left[ \mathcal{L}^\text{t}_s(\theta) \right]
\label{actor loss}
\end{equation}
where the per-state loss is:
\begin{equation}
\mathcal{L}^\text{t}_s(\theta) = -\mathbb{E}_{a \sim \pi^\text{t}_\theta(\cdot|o^\text{t})} \left[Q^\text{t}_\phi(o^\text{t},a) - \alpha \log \pi^\text{t}_\theta(a|o^\text{t}) \right]
\end{equation}

The policy is trained until it achieves a $100\%$ success rate on the training tasks. The combination of the relative pose representation and camera images $\mathcal{I}$ equips the policy with reactive visual-servoing behavior that is robust to control inaccuracies or physical disturbances. However, as we show in Sec.~\ref{sec ablation}, the vision-enabled policy's performance is highly sensitive to the environment's appearance.

\subsection{Vision-free policy training via teacher-student distillation} \label{sec vision-free student distillation}

In order to overcome visual task overfitting while retaining reactive behavior, we distill the vision-enabled teacher $\pi^{\text{t}}$ into a vision-free student $\pi^{\text{s}}$ that observes only the proprioceptive state $o^\text{s} = x^{\text{s}}$. To avoid spatial overfitting, the vision-free policy observes a modified proprioceptive state 
\begin{equation}
    x^\text{s} = \big( T_z(p), \; v, \; f, \; \tau \big).
    \label{modified vision-free proprio}
\end{equation}
where $T_z(\cdot)$ is a homogeneous transformation to a task-relative coordinate system for a given task $z$.

The vision-free policy is trained using the same HIL-RL framework described in Sec.~\ref{sec preliminary}, but with additional distillation terms applied to the actor and critic loss functions.

We encourage the vision-free student's critic $Q^{\text{s}}$ to match the converged vision-enabled teacher's critic $Q^{\text{t}}$ by adding a mean-square error term to its critic loss function \eqref{critic loss}:

\begin{equation}
\begin{aligned}
\mathcal{L}_{Q}^\text{s}(\phi) = \mathbb{E}_{(o,a,r,o') \sim \mathcal{D}^\text{s}_\text{HIL}} [( Q^\text{s}_{\phi}(o^\text{s}, a) - y )^2 \\ + \big(Q^{\text{t}}(o^{\text{t}},a)-Q^{\text{s}}(o^\text{s},a)\big)^2 ],
\label{critic loss}
\end{aligned}
\end{equation}
where:
\begin{equation}
y = r + \gamma \left( \min_{i=1,2} Q^\text{s}_{\bar{\phi}_i}({o^\text{s}}', a') - \alpha \log \pi^\text{s}_\theta(a'|{o^\text{s}}') \right).
\end{equation}

Furthermore, we regularize the vision-free student $\pi^{\text{s}}$ to match the vision-enabled teacher’s actions by adding a Kullback-Leibler (KL) divergence term to its actor loss function:
 \begin{equation}
\mathcal{L}^\text{s}_\pi(\theta) = \mathbb{E}_{o^\text{s} \sim \mathcal{D}^\text{s}_\text{HIL}} \left[ \mathcal{L}_s^\text{s}(\theta) \right]
\label{actor loss g}
\end{equation}
where the per-state loss is:
\begin{equation}
\begin{aligned}
\mathcal{L}_s^\text{s}(\theta) = -\mathbb{E}_{a \sim \pi^\text{t}_\theta(\cdot|o^\text{s})} \big[& Q^\text{s}_\phi(o^\text{s},a) - \alpha \log \pi^\text{s}_\theta(a|o^\text{s}) \\
& - D_{\mathrm{KL}}(\pi^{\text{t}}(a|o) \, \Vert \, \pi^{\text{s}}(a|o^\text{s})) \big]
\end{aligned}
\end{equation}

\subsection{Zero-shot generalization and fine-tuning}\label{sec zeroshot fewshot}

At this point, we can evaluate the zero-shot performance of the trained vision-free policy $\pi^{\text{s}}$ on unseen tasks given an estimated target task pose to obtain the task-relative proprioceptive states \eqref{modified vision-free proprio}. Our method's performance for those unseen tasks is shown in Sec.~\ref{sec exp zeroshot}, alongside a comparison with relevant baselines. 

For more challenging tasks, a few additional training episodes can be performed to fine-tune the vision-free policy by replacing the previously employed vision-enabled teacher with a new one, ${\pi^{\text{t}}}'$, specialized in the new task. We show in our experiments (Sec.~\ref{sec exp fewshot}) that this approach significantly increases task performance.

\section{Experimental Setup}

This section lists the experimental conditions used to evaluate our method, including the hardware, algorithmic details, benchmark tasks, and baseline implementations.

\subsection{Hardware}

Our robotic platform consists of a Franka FR3 robot with two wrist-mounted Intel RealSense D435i cameras and a 3D Space mouse for human teleoperation. An NVIDIA RTX A4000 GPU is used for neural network training.

\subsection{Algorithmic details}

Our policies are executed at 10Hz, with an episode length of 100 steps (10 seconds) during training and 150 steps (15 seconds) during evaluation. Training lasts 50 minutes. Of those, 40 minutes are used to train the vision-enabled teacher, and 10 minutes to train the vision-free student. We fill the human data buffer from Sec.~\ref{sec rlpd} with 5 demonstrations per task before training and apply interventions to promote task success during the rollouts. 
Table~\ref{table_rlparams} lists the observation space, action space, and network architecture. We employ separate MLP networks for the actor and the critics, with a shared observation encoder backbone. Visual input is processed by a frozen, pretrained ResNet-10 encoder; proprioception is encoded by a projection layer; and both are concatenated. Actor, critic, and projection layers use LayerNorm and Tanh activations. 

\begin{table}[h]
\caption{RL specifications}
\vspace{-0.3cm}
\label{table_rlparams}
\begin{center}
\begin{tabular}{l c}
\toprule
 \textbf{Algorithm Settings} &  \textbf {Values }\\
\toprule
 Visual observations &  Two $128\times128$ RGB views \\
 Action space (translation) & $3 \times 3 \times 3$ cm along $x, y, z$ \\
  Action space (rotation) & $10 \times 10 \times 10^{\circ}$ about $x, y, z$ \\
 
 Image encoder &  Pretrained ResNet-10 (frozen) \\

  Proprioception encoder &   FC $64$ \\
  Actor network & MLP $[256,256]$ \\
  Critic network & MLP $[256,256]$ \\
  \bottomrule
\end{tabular}
\end{center}
\vspace{-0.3cm}
\end{table}

\subsection{Task Benchmark}  \label{sec benchmark}
\begin{figure}[htb]
  \centering
  \begin{subfigure}[b]{0.4\linewidth}
    \centering
    \includegraphics[width=\linewidth]{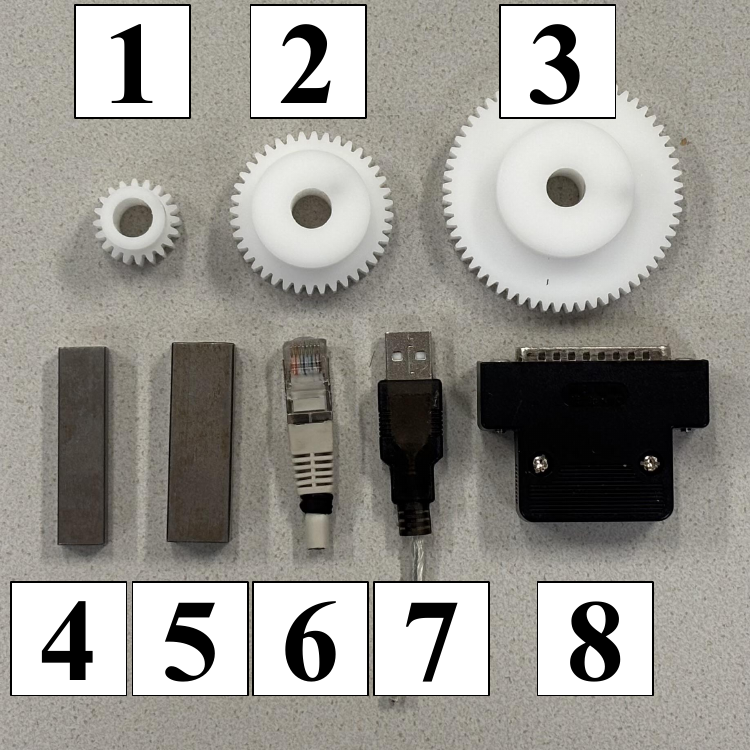}
    \caption{All connectors.\newline}
    \label{fig:plugs}
  \end{subfigure}
  \begin{subfigure}[b]{0.4\linewidth}
    \centering
    \includegraphics[width=\linewidth]{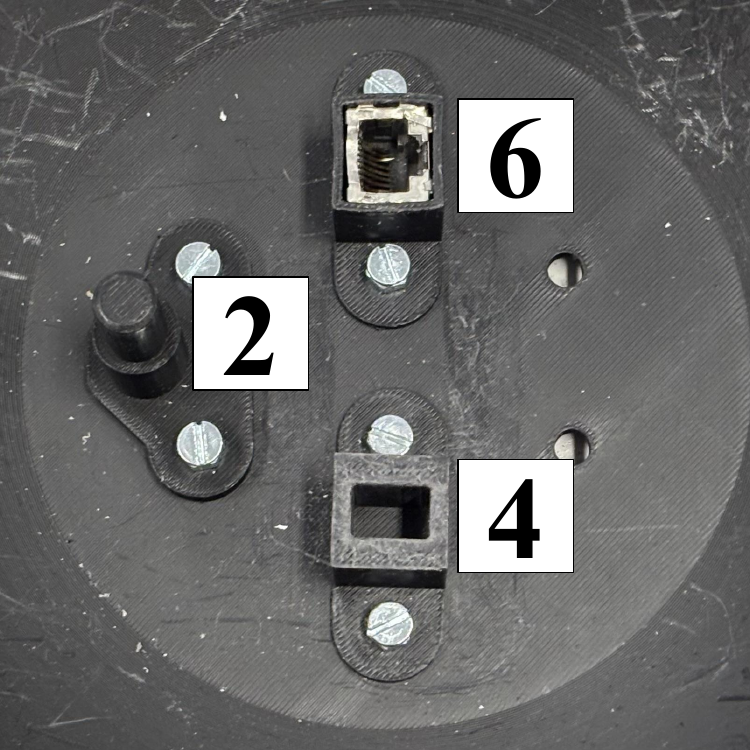}
    \caption{Training tasks.\newline}
    \label{fig:train_tasks}
  \end{subfigure}
  
  \begin{subfigure}[b]{0.5\linewidth}
    \centering
    \includegraphics[width=\linewidth]{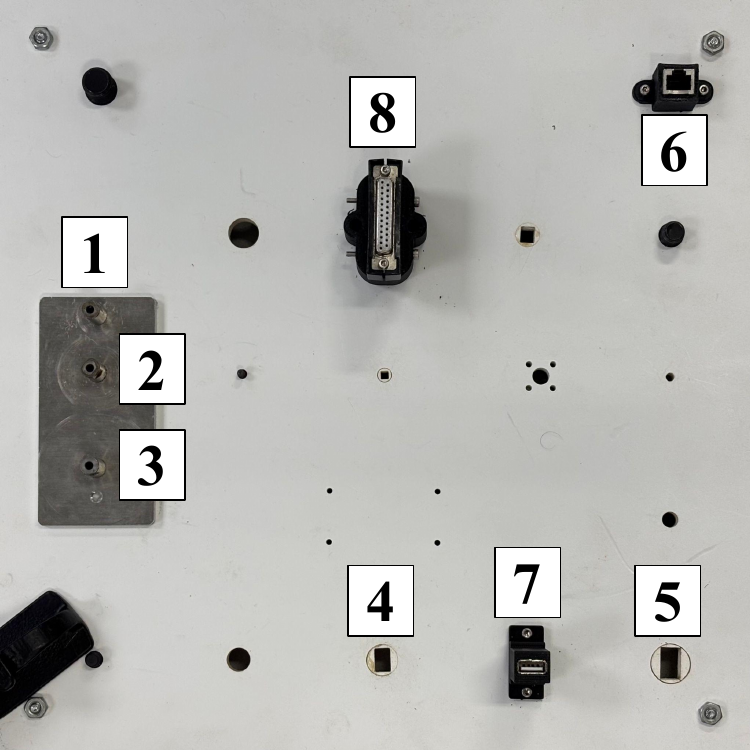}    
    \caption{Test tasks - NIST Assembly Board I (unseen in training).}
    \label{fig:eval_tasks}
  \end{subfigure}
  \caption{Benchmark tasks: 1. S gear, 2. M gear, 3. L gear, 4. M peg, 5. L peg, 6. ethernet, 7. USB, 8. DSUB. (S, M, L stand for small, medium, large.)}
   \vspace{-0.4cm}
  \label{fig:training_eval_setup}
\end{figure}

The benchmark tasks are based on the NIST Assembly Board I (Fig.~\ref{fig:training_eval_setup}), a standardized platform for evaluating manipulation performance across a range of precision assembly tasks. 

Policies for our method and all baselines are trained on the 3 tasks in Fig.~\ref{fig:train_tasks}, using a 3D-printed training environment. Each of the tasks has a particular contact pattern: medium peg is rectangular and stiff, ethernet is asymmetric and soft, and medium gear is round and constrained in the z-axis (i.e., if the hole is not aligned, the gear is kept on top of the socket, as opposed to what happens in Fig.~\ref{fig:usb}). Policy training was sequentially conducted across multiple tasks. Specifically, 5 training episodes (trials) are conducted for the first task, followed by 5 for the second and 5 for the third, after which the training cycle returns to the first task. This loop is repeated until the policy converges.

Evaluation is conducted on the eight tasks in Fig.~\ref{fig:eval_tasks}, now on a replica of the  NIST Assembly Board I. While connectors 2, 4, and 6 (Fig.~\ref{fig:plugs}) appear in both training and evaluation, the corresponding tasks (sockets) remain distinct. 

For the task-relative pose representation $T_z(p)$ in Sec.~\ref{sec vision-free student distillation} used by our method and all the vision-free baselines, each task's pose is estimated by kinesthetically guiding the robot through the task and recording the final pose.



\section{Experimental Results}

Our experiments aim to answer the following three main questions: 

\begin{itemize}
\item How does our method compare to baselines in terms of robustness and generalization to unseen tasks?
\item How do different sensory input modalities contribute to policy performance?
\item Can fine-tuning with distillation achieve complete success on challenging tasks?
\end{itemize}

\subsection{Robustness and generalization compared to baselines} \label{sec exp zeroshot}
We compare our method to three baselines:

\subsubsection{HIL-SERL \cite{hilserl}} \label{sec hil rl proprio baseline} human-in-the-loop RL setting employed in our method. We try two variants: \textit{VPTW}, using vision, pose, twist, and wrench as input, which is equivalent to the vision-enabled teacher of our method, and \textit{PTW}, using pose, twist, and wrench as input, which is similar to the vision-free student of our method, but trained from scratch, without our proposed distillation. \textit{VPTW} is trained for 40 minutes, and PTW for 50 minutes. 

\subsubsection{Dynamic Motion Primitive (DMP) \cite{DMP}} \label{sec dmp baseline} 
represents motions through a system of nonlinear differential equations that captures the essential dynamics of a demonstrated motion while enabling adaptation to different initial and goal states. We use a basic open-loop DMP learned from a single kinesthetic demonstration of the insertion task with the medium peg. Uses only pose input, and does not require training.

\subsubsection{DMP + Residual RL \cite{residualfromdemo}} \label{sec residual rl baseline} augments the DMP baseline with a residual SAC agent trained for 50 minutes, whose corrective actions are added to the DMP trajectory at each timestep. This combines demonstration-driven motion with adaptive RL-based correction.

\begin{figure}[!htb]
  \centering
   \begin{subfigure}[b]{0.44\linewidth}
    \centering
    \begin{tikzpicture}
        \node[inner sep=0pt] (img) {\includegraphics[width=\linewidth]{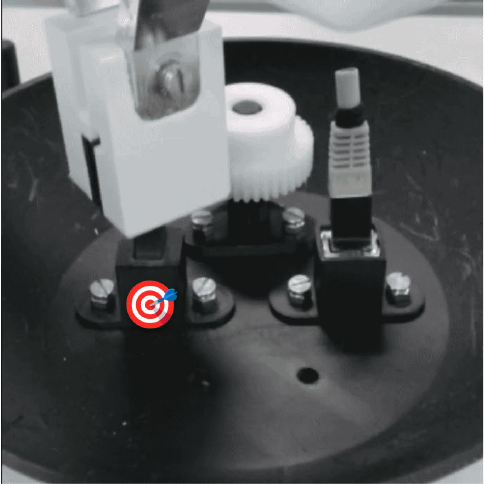}};
        \fill[white, opacity=0.1] (img.south west) rectangle (img.north east);
    \end{tikzpicture}
     \caption{Normal conditions}
     \label{training conditions}
    \end{subfigure}
    \begin{subfigure}[b]{0.44\linewidth}
    \centering
    \begin{tikzpicture}
        \node[inner sep=0pt] (img) {\includegraphics[width=\linewidth]{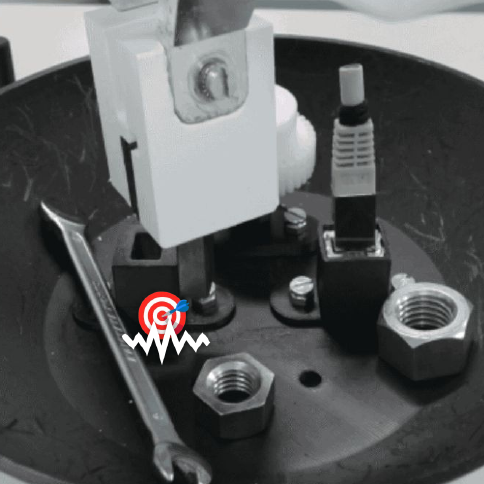}};
        \fill[white, opacity=.10] (img.south west) rectangle (img.north east);
    \end{tikzpicture}
    \caption{Disturbed conditions}
    \label{disturbed conditions}
    \end{subfigure}
  \caption{Training tasks in (a) normal conditions and (b) disturbed conditions. The disturbed case contains visual distractors (screws and tool) and random target pose noise ranging from 5-8mm in translation (xyz) and 1-3° in rotation around the z-axis, which perturbs the task-relative pose representation in \eqref{modified vision-free proprio}.}
   \vspace{-0.2cm}
  \label{visual success fail}
\end{figure}

All methods are evaluated on three task categories: the training tasks (Fig.~\ref{fig:train_tasks}), a disturbed version of the training tasks that introduces visual distractors and target pose uncertainty (Fig.~\ref{visual success fail}), and unseen tasks with new geometries (Fig.~\ref{fig:eval_tasks}).

The results are summarized in Table~\ref{table_unseenbaseline}. HIL-SERL with visual, pose, twist, and wrench input (\textit{VPTW}) achieves near-perfect success on training tasks, yet degrades substantially under disturbed conditions and collapses entirely on unseen tasks, yielding the lowest overall success rate. The peg task is particularly affected under disturbed conditions, likely due to its higher difficulty or the concentration of visual distractors in its vicinity (Fig.~\ref{disturbed conditions}). These results confirm our core hypothesis: vision-based policies overfit to training-specific visual features and fail to generalize. This brittleness motivates our distillation approach, which leverages visual information during training but discards it at deployment.

DMP shows similarly poor overall performance, confirming that fixed motion primitives lack the adaptability required for robust contact-rich manipulation. HIL-SERL with pose, twist, and wrench input (\textit{PTW}) shows moderate improvement due to RL's closed-loop nature, but still struggles to learn effective insertion strategies from low-dimensional sensory input alone.

Residual RL performs considerably better, as the base policy acts as an attractor towards the sockets while the learned residual handles contact-rich behaviors such as wiggling and alignment correction. This baseline is comparable to our method in geometrically simpler tasks, such as medium and large pegs and gears.

\begin{figure}[!htb]
  \centering
  \begin{subfigure}[b]{0.18\linewidth}
    \centering
    \begin{tikzpicture}
        \node[inner sep=0pt] (img) {\includegraphics[trim=1200pt 820pt 1200pt 600pt, clip, width=\linewidth]{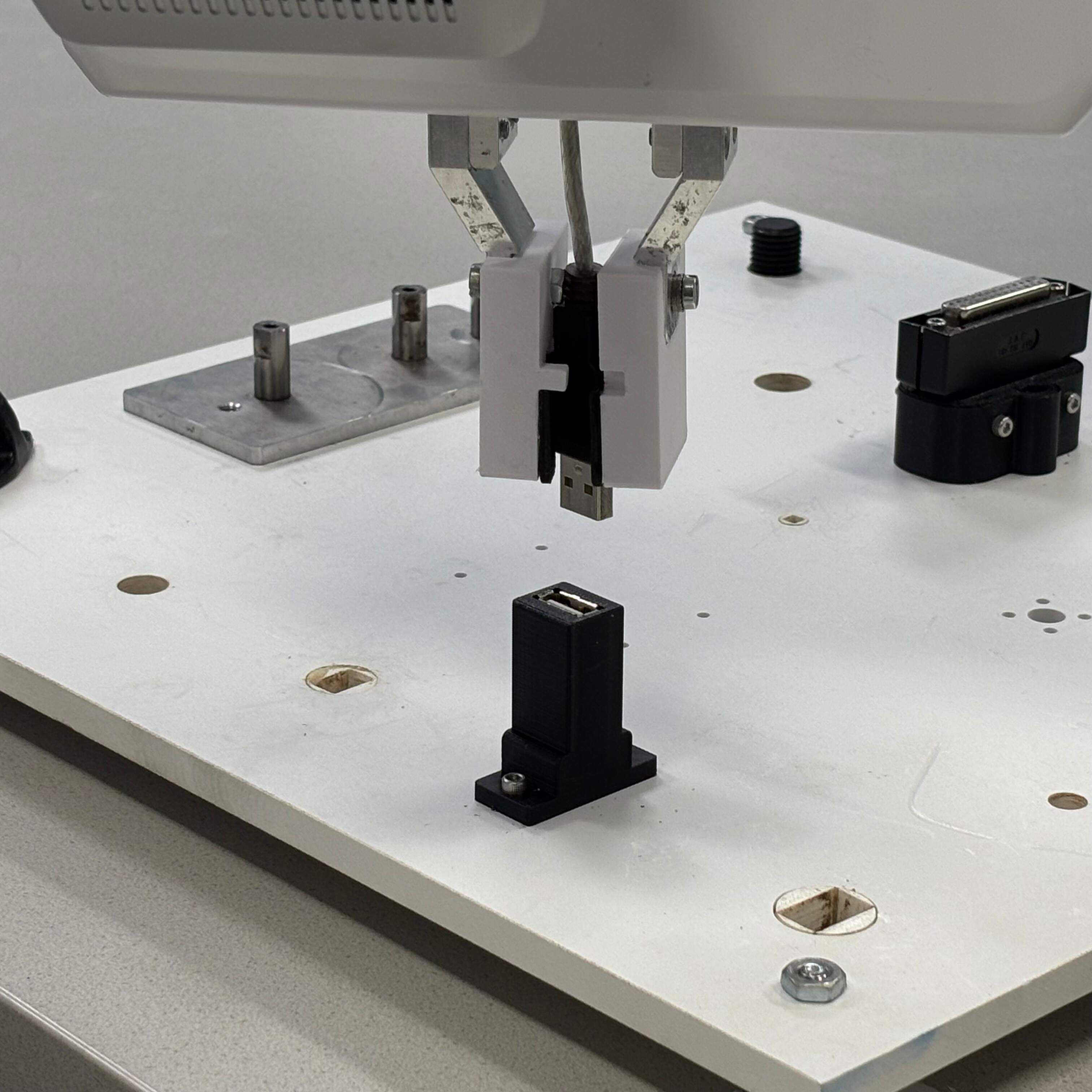}};
        \fill[white, opacity=0.1] (img.south west) rectangle (img.north east);
    \end{tikzpicture}
    \caption{Initialize}
  \end{subfigure}
  \hfill
  \begin{subfigure}[b]{0.18\linewidth}
    \centering
    \begin{tikzpicture}
        \node[inner sep=0pt] (img) {\includegraphics[trim=1200pt 820pt 1200pt 600pt, clip, width=\linewidth]{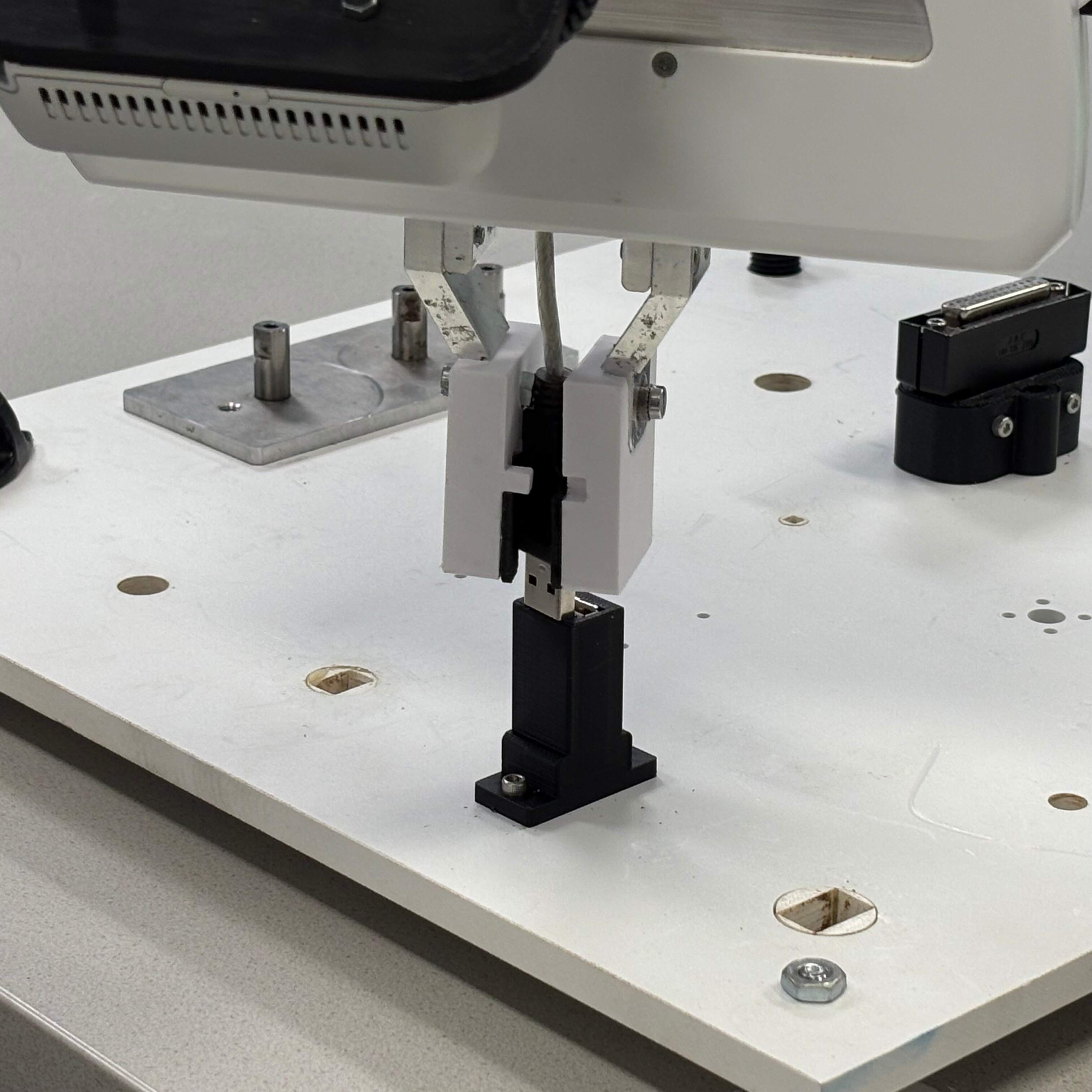}};
        \fill[white, opacity=0.1] (img.south west) rectangle (img.north east);
    \end{tikzpicture}
    \caption{Miss}
  \end{subfigure}
  \hfill
  \begin{subfigure}[b]{0.18\linewidth}
    \centering
    \begin{tikzpicture}
        \node[inner sep=0pt] (img) {\includegraphics[trim=1200pt 820pt 1200pt 600pt, clip, width=\linewidth]{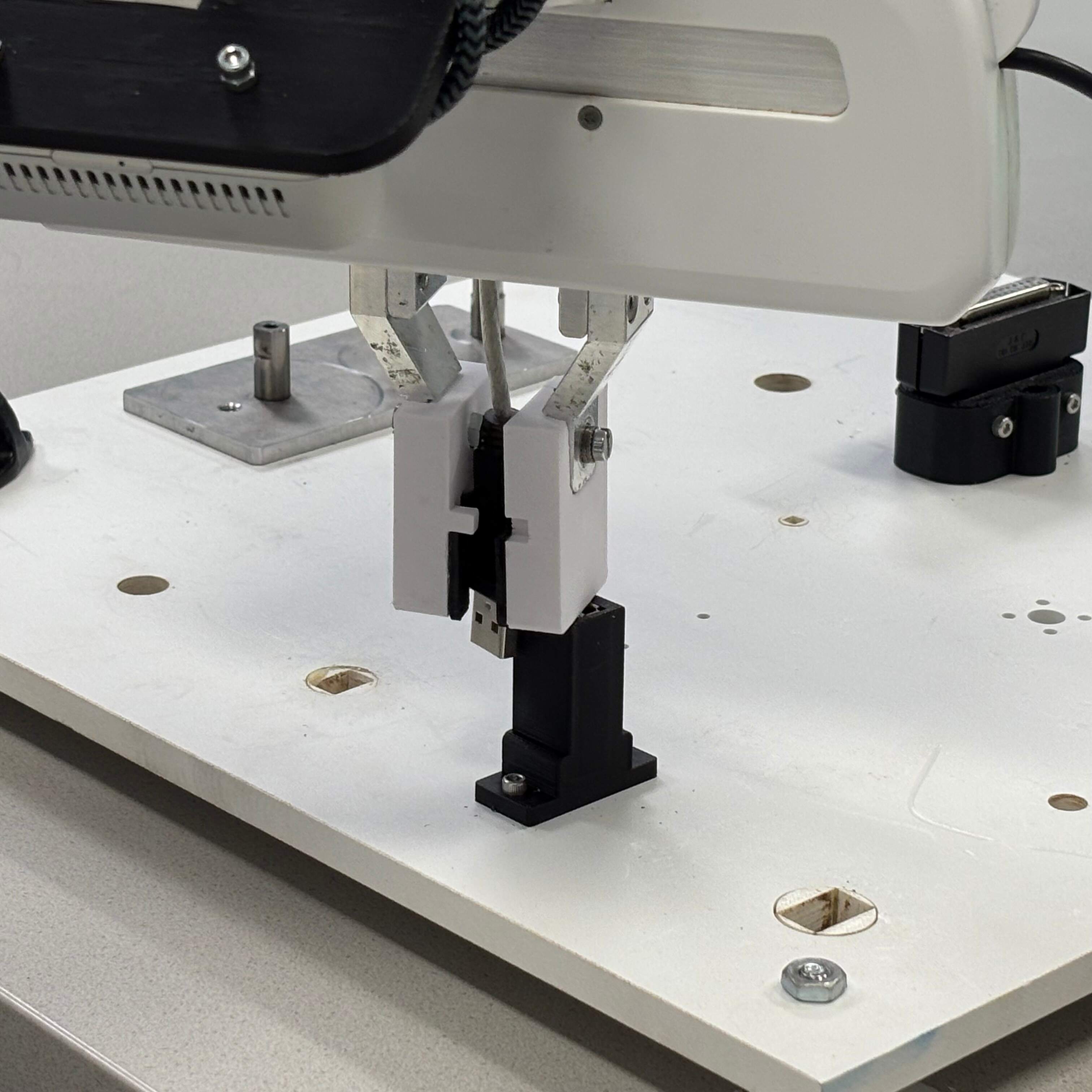}};
        \fill[white, opacity=0.1] (img.south west) rectangle (img.north east);
    \end{tikzpicture}
    \caption{Slip}
  \end{subfigure}
  \hfill
  \begin{subfigure}[b]{0.18\linewidth}
    \centering
   \begin{tikzpicture}
        \node[inner sep=0pt] (img) {\includegraphics[trim=1200pt 820pt 1200pt 600pt, clip, width=\linewidth]{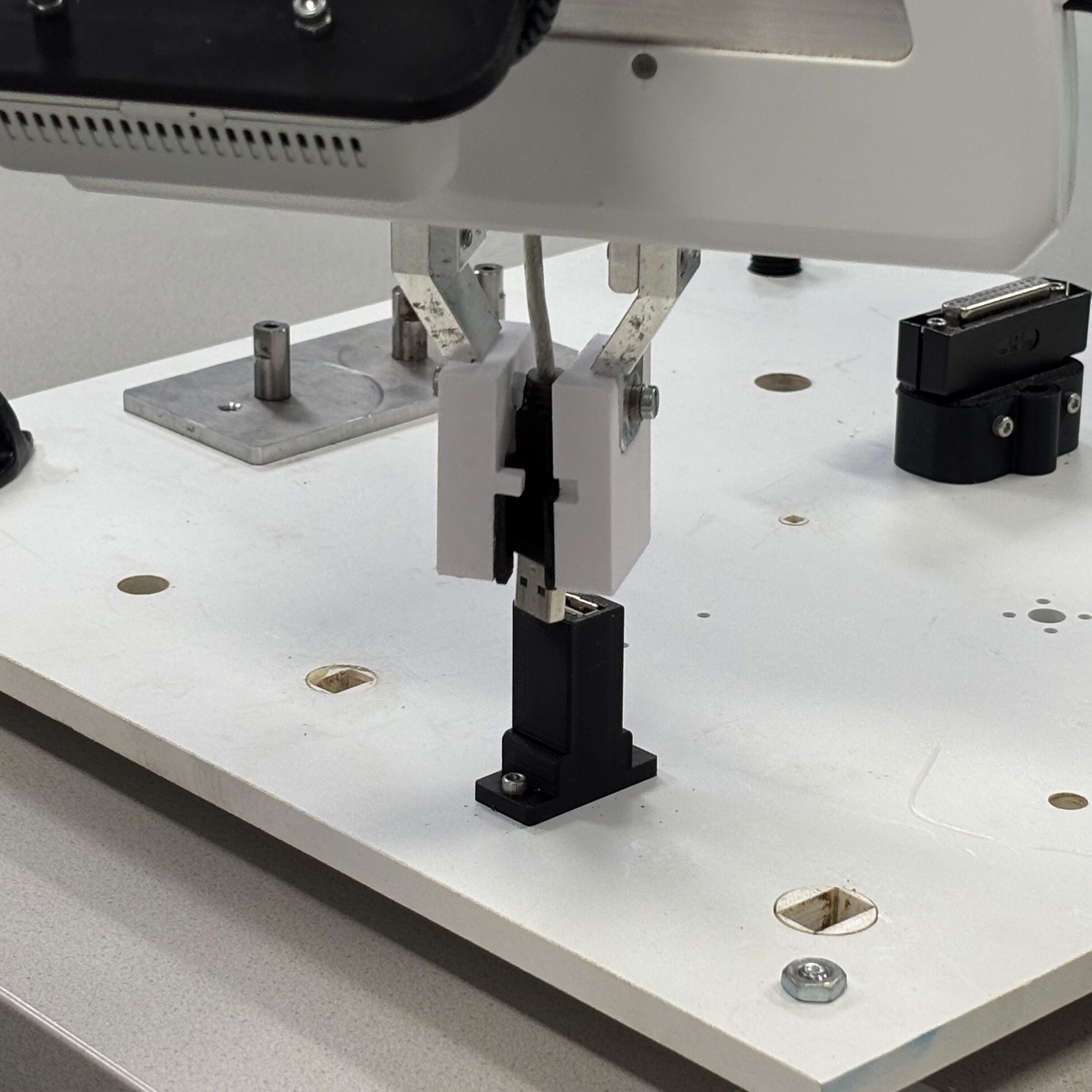}};
        \fill[white, opacity=0.1] (img.south west) rectangle (img.north east);
    \end{tikzpicture}
    \caption{Retry}
  \end{subfigure}
  \hfill
  \begin{subfigure}[b]{0.18\linewidth}
    \centering
    \begin{tikzpicture}
        \node[inner sep=0pt] (img) {\includegraphics[trim=1200pt 820pt 1200pt 600pt, clip, width=\linewidth]{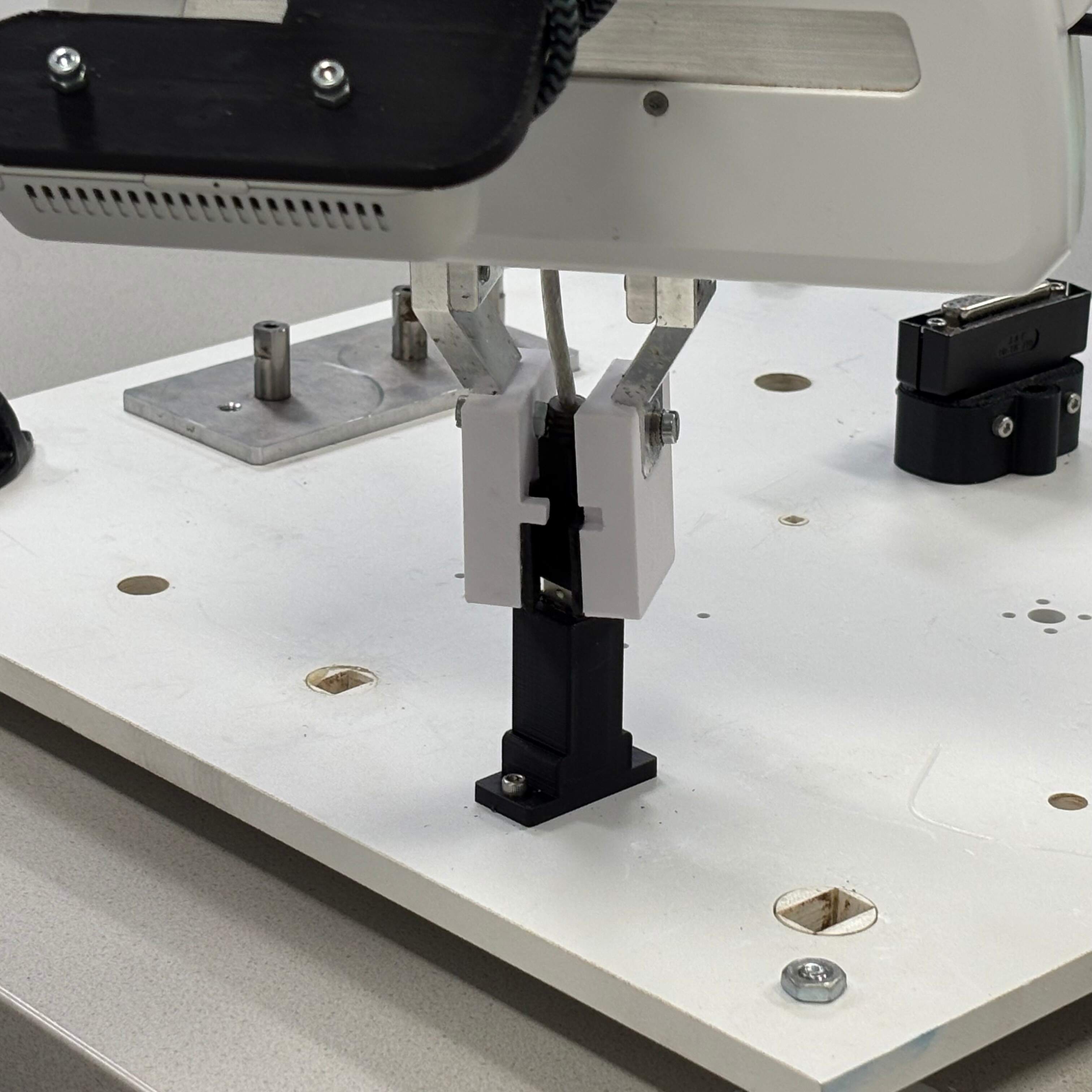}};
        \fill[white, opacity=0.1] (img.south west) rectangle (img.north east);
    \end{tikzpicture}
    \caption{Inserted}
  \end{subfigure}
  \hfill
  \caption{Zero-shot USB insertion by our method \textit{VE2VF}, showing a reactive recovery behavior: the robot misses the socket, slips off, retries, and successfully inserts.}
  \label{fig:usb}
  \vspace{-0.3cm}
\end{figure}

Our method \textit{VE2VF} achieves the highest overall success rate at 95.0\%, with notably superior robustness under disturbed conditions and generalization to unseen tasks. The pronounced advantage in USB insertion reflects a qualitative recovery behavior when the robot slips off the socket surface, observed only in our method (Fig.~\ref{fig:usb}). We attribute these recovery strategies to the reactive vision-enabled teacher, whose visual-servoing behavior is effectively distilled into the proprioceptive student. This can be better visualized in our Supplementary Video.

\begin{table*}[h]
\caption{Success rates across training, disturbed, and unseen insertion tasks for our method and baselines. Input modalities: V=vision, P=pose, T=twist, W=wrench.}
\vspace{-0.3cm}
\label{table_unseenbaseline}
\begin{center}
\setlength{\tabcolsep}{2pt}
\begin{tabular}{l l c c c c c c c c c c c c c c c}
\toprule
 & & \multicolumn{3}{c}{Training tasks (Fig.~\ref{fig:train_tasks})} & \multicolumn{3}{c}{Disturbed tasks (Fig.~\ref{disturbed conditions})} & \multicolumn{8}{c}{Unseen tasks (Fig.~\ref{fig:eval_tasks})} & \multirow{2}{*}[-2pt]{Overall} \\
 \cmidrule(lr){3-5} \cmidrule(lr){6-8} \cmidrule(lr){9-16}
 & Input & M Peg & Ethernet & M Gear & M Peg & Ethernet & M Gear & M Peg & L Peg & Ethernet & USB & DSUB & S Gear & M Gear & L Gear & \\
\midrule
HIL-SERL \cite{hilserl} & \textit{VPTW} &  \textbf{10/10} &  \textbf{10/10} &  \textbf{10/10} & 0/10 &  \textbf{10/10} & 8/10 & 0/10 & 0/10 & 0/10 & 0/10 & 0/10 & 0/10 & 0/10 & 0/10 & 34.3\% \\
HIL-SERL \cite{hilserl} & \textit{PTW} & 7/10 & 6/10 & 9/10 & 5/10 & 4/10 & 8/10 & 6/10 & 8/10 & 0/10 & 5/10 & 2/10 & 1/10 & 4/10 & 1/10 & 47.1\% \\
DMP \cite{DMP} & \textit{P} & 9/10 & 5/10 & 9/10 & 4/10 & 2/10 & 2/10 & 4/10 & 3/10 & 5/10 & 0/10 & 0/10 & 0/10 & 4/10 & 3/10 & 35.7\% \\
Residual RL \cite{residualfromdemo} & \textit{PTW} & 9/10 & \textbf{10/10} & \textbf{10/10} & 8/10 & 9/10 & \textbf{10/10} & 9/10 & 9/10 & 9/10 & 6/10 & \textbf{6/10} & 7/10 & \textbf{10/10} & 8/10 & 85.7\% \\
\textbf{\textit{VE2VF}} & \textit{PTW} & \textbf{10/10} & \textbf{10/10} & \textbf{10/10} & \textbf{10/10} & \textbf{10/10} & \textbf{10/10} & \textbf{10/10} & \textbf{10/10} & \textbf{10/10} & \textbf{10/10} & 5/10 & \textbf{9/10} & \textbf{10/10} & \textbf{9/10} & \textbf{95.0\%} \\
\bottomrule
\end{tabular}
\vspace{-0.3cm}
\end{center}
\end{table*}

\subsection{Input modalities ablation study}
\label{sec ablation}

Having established our method's advantage over baselines, we now analyze how individual sensory modalities contribute to performance. All policies are evaluated under both normal and disturbed conditions (Fig.~\ref{visual success fail}). Results are shown in Table~\ref{table_modalities}.

We first compare vision-enabled policies. Note that \textit{VPTW} is equivalent to our vision-enabled teacher and to the HIL-SERL \textit{VPTW} baseline in Section~\ref{sec exp zeroshot}. A vision-only policy (\textit{V}) matches \textit{VPTW} under training conditions, but deteriorates more severely under disturbance, suggesting that the additional proprioceptive inputs in \textit{VPTW} partially compensate when visual features become unreliable.

Similarly, \textit{PTW} corresponds to the HIL-SERL \textit{PTW} baseline, a proprioceptive policy trained from scratch without distillation. As shown in Section~\ref{sec exp zeroshot}, it fails to reach full task completion, underscoring the importance of visual input during training.

We then compare vision-free student policies, denoted by the prefix \textit{s}, distilled for 10 minutes from a \textit{VPTW} teacher trained for 40 minutes. \textit{sPTW} uses pose, twist, and wrench, \textit{sP} uses only pose, and \textit{sTW} uses twist and wrench with a history of the past eight observations. The results reveal two key findings: first, pose information is important, as \textit{sTW} performs worse than the other two despite having a longer observation horizon; second, combining all proprioceptive modalities improves robustness, as \textit{sPTW} maintains perfect performance under disturbance while \textit{sP} degrades significantly, since it relies more heavily on the pose information affected by the target pose noise.

\begin{figure}[!tbp]
  \centering
  \includegraphics[trim=0pt 9pt 0pt 9pt, clip, width=0.9\linewidth]{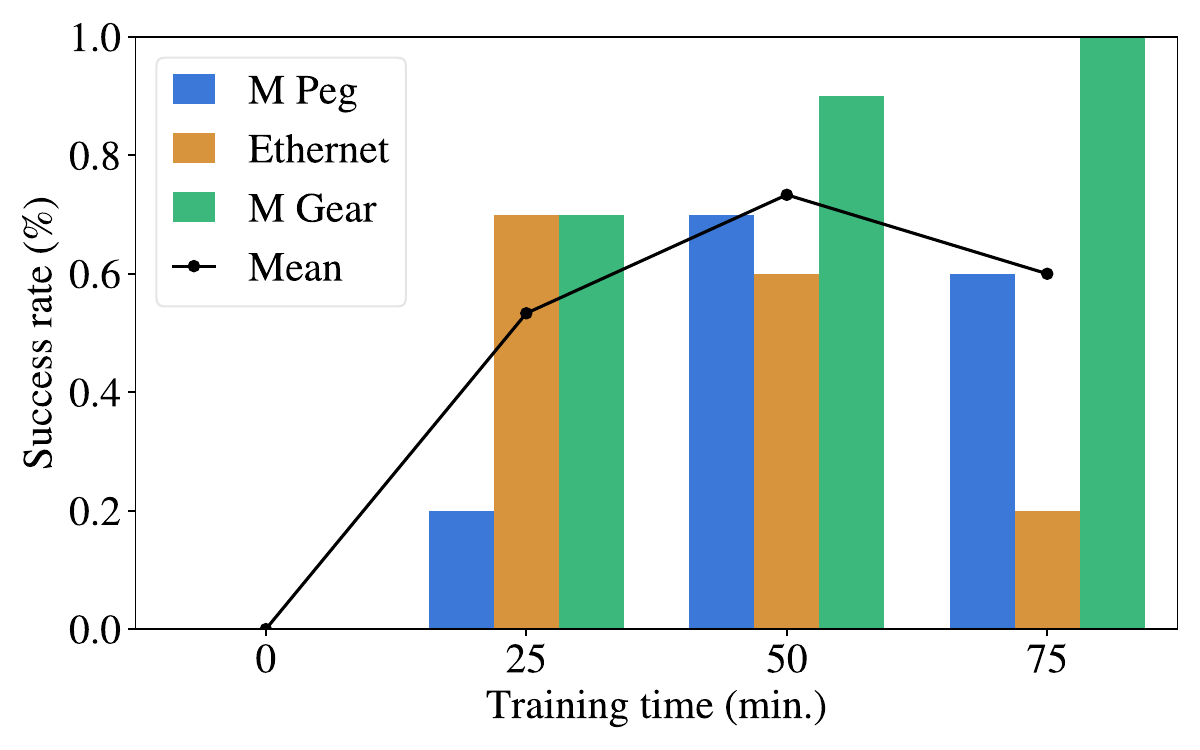}
  \vspace{-0.2cm}
\caption{Training progression of the vision-free policy \textit{PTW}, trained from scratch without distillation. Performance plateaus after 50 minutes, with longer training leading to overfitting on the gear task.}
   \vspace{-0.4cm}
  \label{propro training barplots}
\end{figure}

Finally, we investigate whether \textit{PTW} simply needs more training time to match our distilled policy. Figure~\ref{propro training barplots} shows success rates for training durations up to 75 minutes ($1.5\times$ our standard). Performance does not improve beyond 50 minutes; rather, the policy overfits to the gear task, presumably the easiest of the three. This confirms that distillation from a vision-enabled teacher is essential, as the proprioceptive input alone does not provide sufficient context for the RL agent to develop a generalized insertion strategy from scratch.

\begin{table}[h]
\caption{Input modalities ablation on training tasks under normal and disturbed conditions. Input modalities: V=vision, P=pose, T=twist, W=wrench. The prefix \textit{s} denotes a student policy distilled for 10 minutes from a \textit{VPTW} teacher trained for 40 minutes. Disturbance consists of visual distractors for policies using vision, or target pose uncertainty for vision-free policies.}
\vspace{-0.3cm}
\label{table_modalities}
\begin{center}
\setlength{\tabcolsep}{2pt}
\begin{tabular}{l c c c c c c c}
\toprule
& & \multicolumn{3}{c}{Training tasks (Fig.\ref{fig:train_tasks})} & \multicolumn{3}{c}{Disturbed tasks (Fig.\ref{disturbed conditions})} \\
 \cmidrule(lr){3-5} \cmidrule(lr){6-8}
 & \multirow{-2}{*}{\begin{tabular}[c]{@{}c@{}}Training\\time (min)\end{tabular}} & M Peg & Ethernet & M Gear & M Peg & M Ethernet & M Gear \\
\midrule

\textit{VPTW} & 40 & \textbf{10/10} & \textbf{10/10} & \textbf{10/10} & 0/10 & \textbf{10/10} & 8/10 \\
\textit{V} & 40 & \textbf{10/10} & \textbf{10/10} & \textbf{10/10}  & 0/10 & 6/10 & 3/10 \\
\addlinespace[1pt]
\arrayrulecolor{black}
\specialrule{0.3pt}{0pt}{0pt}
\arrayrulecolor{black}
\addlinespace[2pt]
\textit{PTW} & 50 & 7/10 & 6/10 & 9/10 &  5/10 & 4/10 & 8/10  \\
\addlinespace[1pt]
\arrayrulecolor{black}
\specialrule{0.3pt}{0pt}{0pt}
\arrayrulecolor{black}
\addlinespace[2pt]
\textit{sPTW} & 40+10 & \textbf{10/10} & \textbf{10/10} & \textbf{10/10}  & \textbf{10/10} & \textbf{10/10} & \textbf{10/10}   \\
\textit{sP} & 40+10 & \textbf{10/10} & \textbf{10/10} & \textbf{10/10}  & 2/10 & 7/10 & 8/10  \\
\textit{sTW} & 40+10 & 4/10 & 7/10 & \textbf{10/10} & 4/10 & 7/10 & \textbf{10/10}  \\
\bottomrule
\end{tabular}
\vspace{-0.3cm}
\end{center}
\end{table}

\subsection{Fine-tuning for challenging tasks}
\label{sec exp fewshot}
\begin{figure}[!tbp]
  \centering
  \begin{subfigure}[b]{0.18\linewidth}
    \centering
    \begin{tikzpicture}
        \node[inner sep=0pt] (img) {\includegraphics[trim=1200pt 850pt 1200pt 600pt, clip, width=\linewidth]{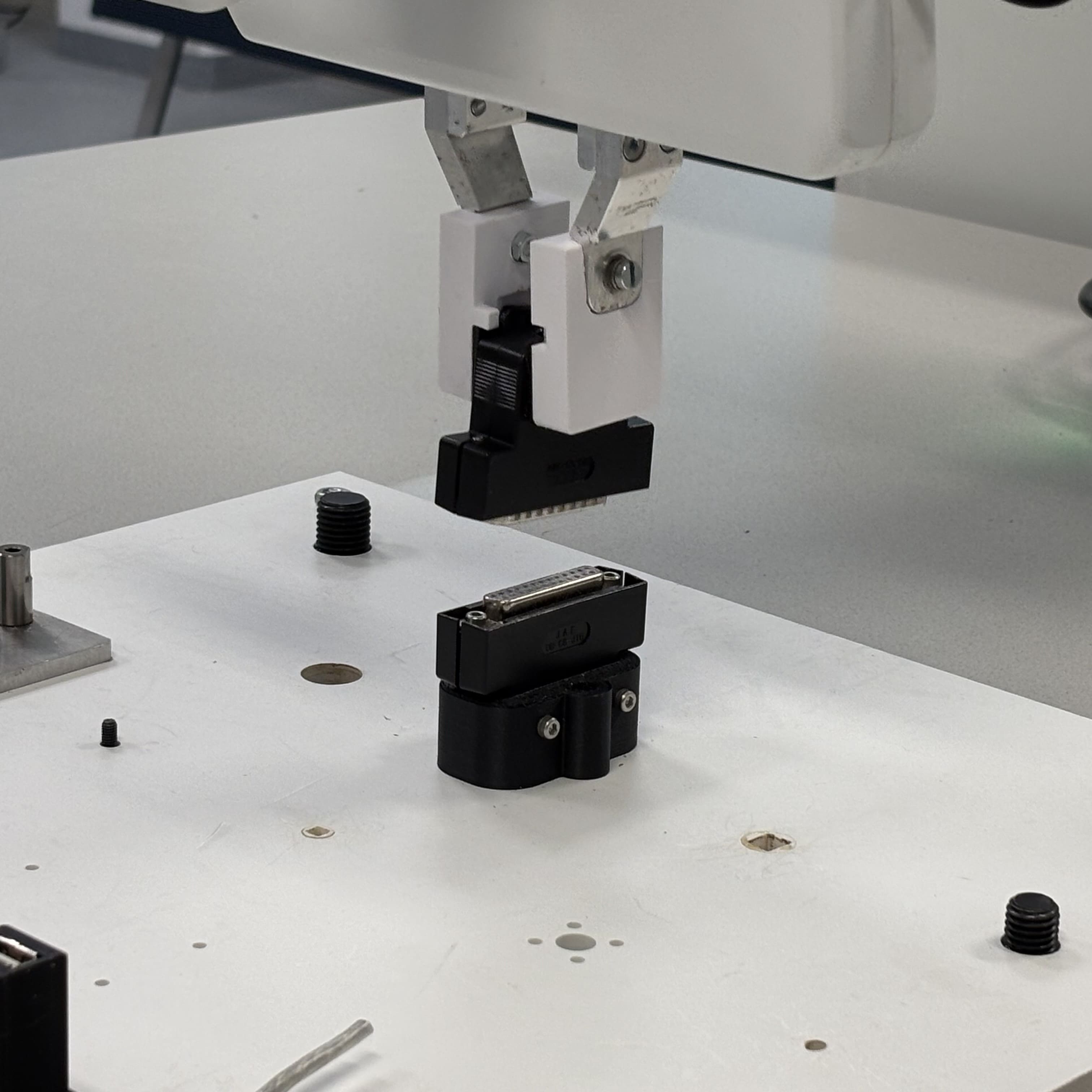}};
        \fill[white, opacity=0.1] (img.south west) rectangle (img.north east);
    \end{tikzpicture}
    \caption{Initialize}
  \end{subfigure}
  \hfill
  \begin{subfigure}[b]{0.18\linewidth}
    \centering
    \begin{tikzpicture}
        \node[inner sep=0pt] (img) {\includegraphics[trim=1200pt 850pt 1200pt 600pt, clip, width=\linewidth]{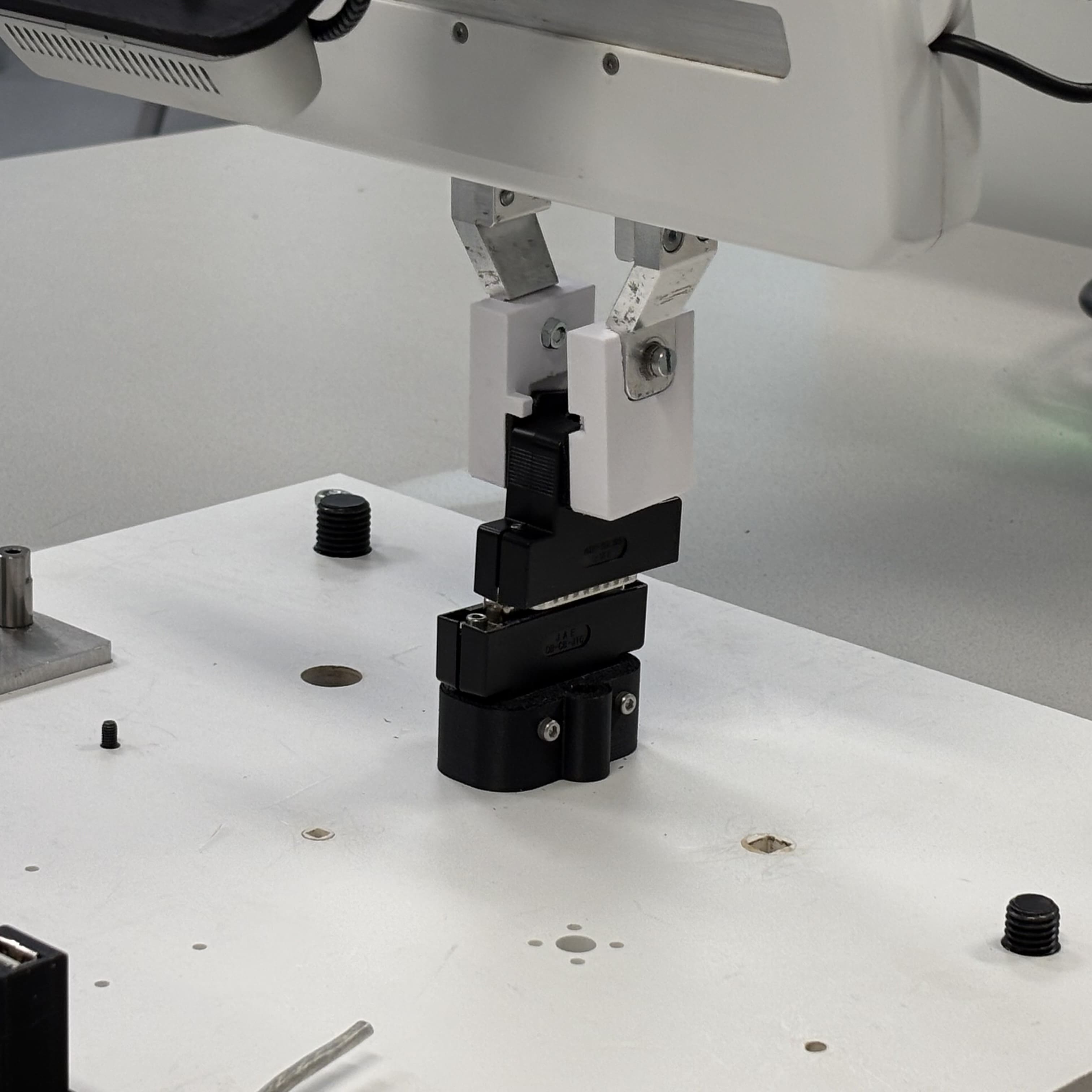}};
        \fill[white, opacity=0.1] (img.south west) rectangle (img.north east);
    \end{tikzpicture}
    \caption{Miss}
  \end{subfigure}
  \hfill
  \begin{subfigure}[b]{0.18\linewidth}
    \centering
    \begin{tikzpicture}
        \node[inner sep=0pt] (img) {\includegraphics[trim=1200pt 850pt 1200pt 600pt, clip, width=\linewidth]{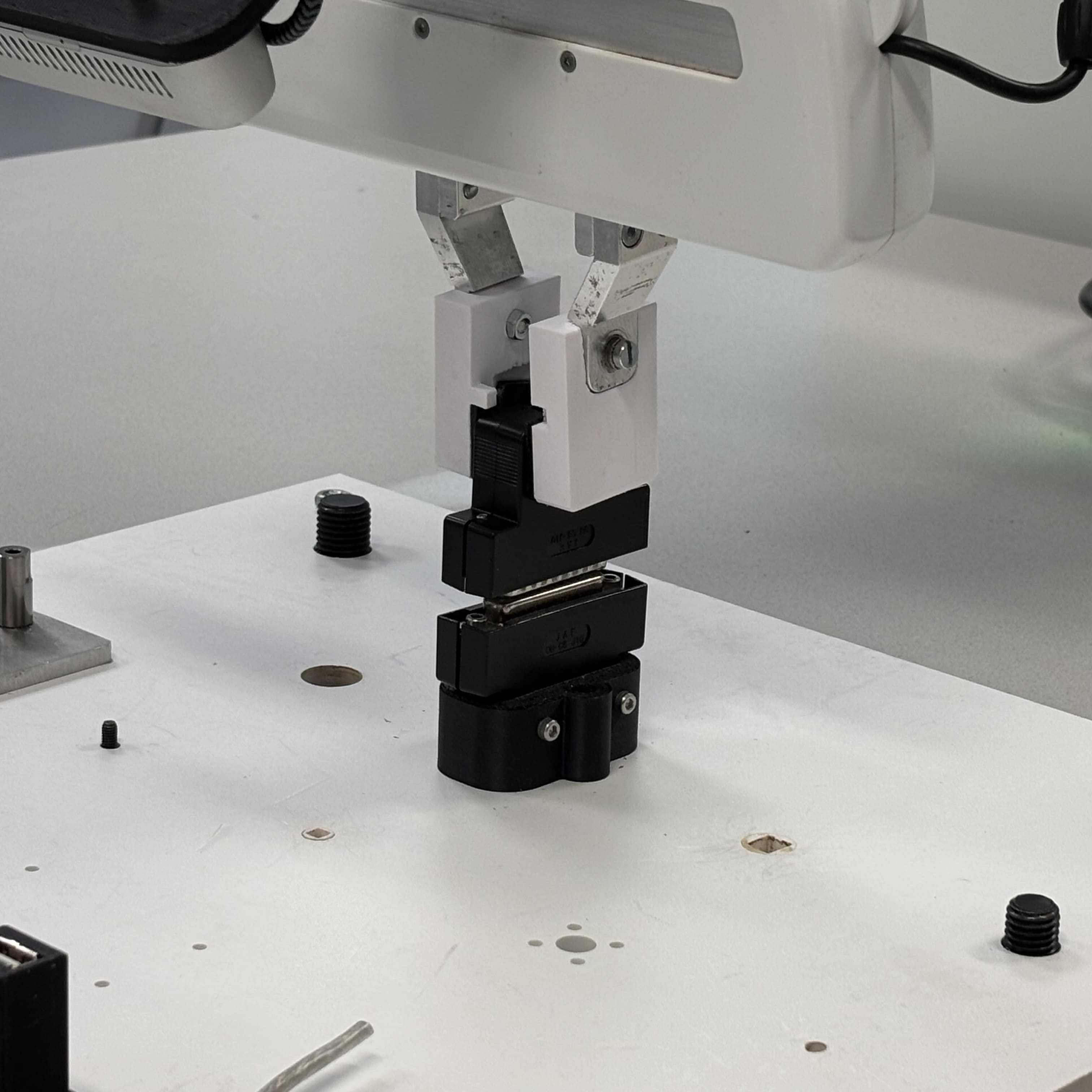}};
        \fill[white, opacity=0.1] (img.south west) rectangle (img.north east);
    \end{tikzpicture}
    \caption{Retry}
  \end{subfigure}
  \hfill
  \begin{subfigure}[b]{0.18\linewidth}
    \centering
   \begin{tikzpicture}
        \node[inner sep=0pt] (img) {\includegraphics[trim=1200pt 850pt 1200pt 600pt, clip, width=\linewidth]{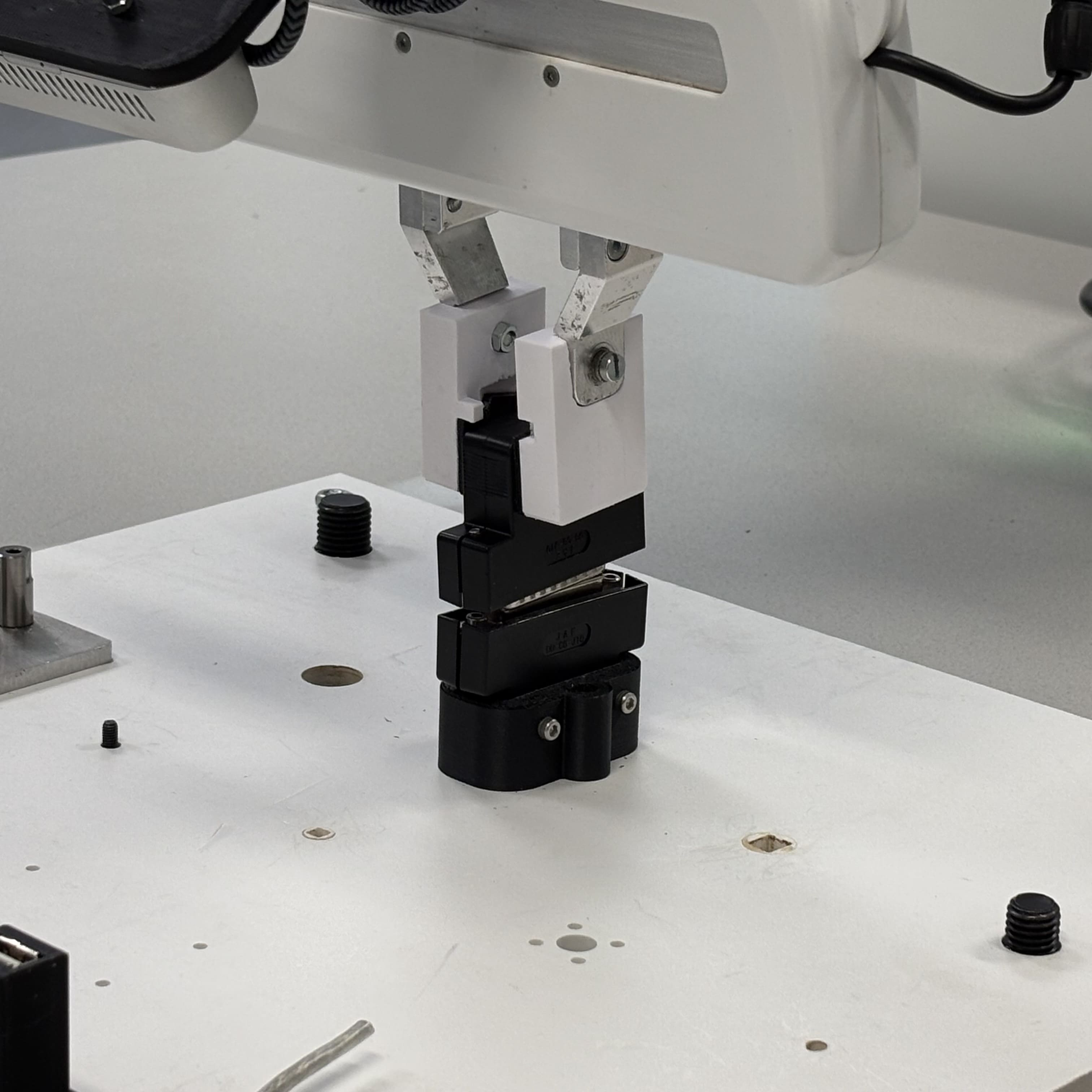}};
        \fill[white, opacity=0.1] (img.south west) rectangle (img.north east);
    \end{tikzpicture}
    \caption{Align}
  \end{subfigure}
  \hfill
  \begin{subfigure}[b]{0.18\linewidth}
    \centering
    \begin{tikzpicture}
        \node[inner sep=0pt] (img) {\includegraphics[trim=1200pt 850pt 1200pt 600pt, clip, width=\linewidth]{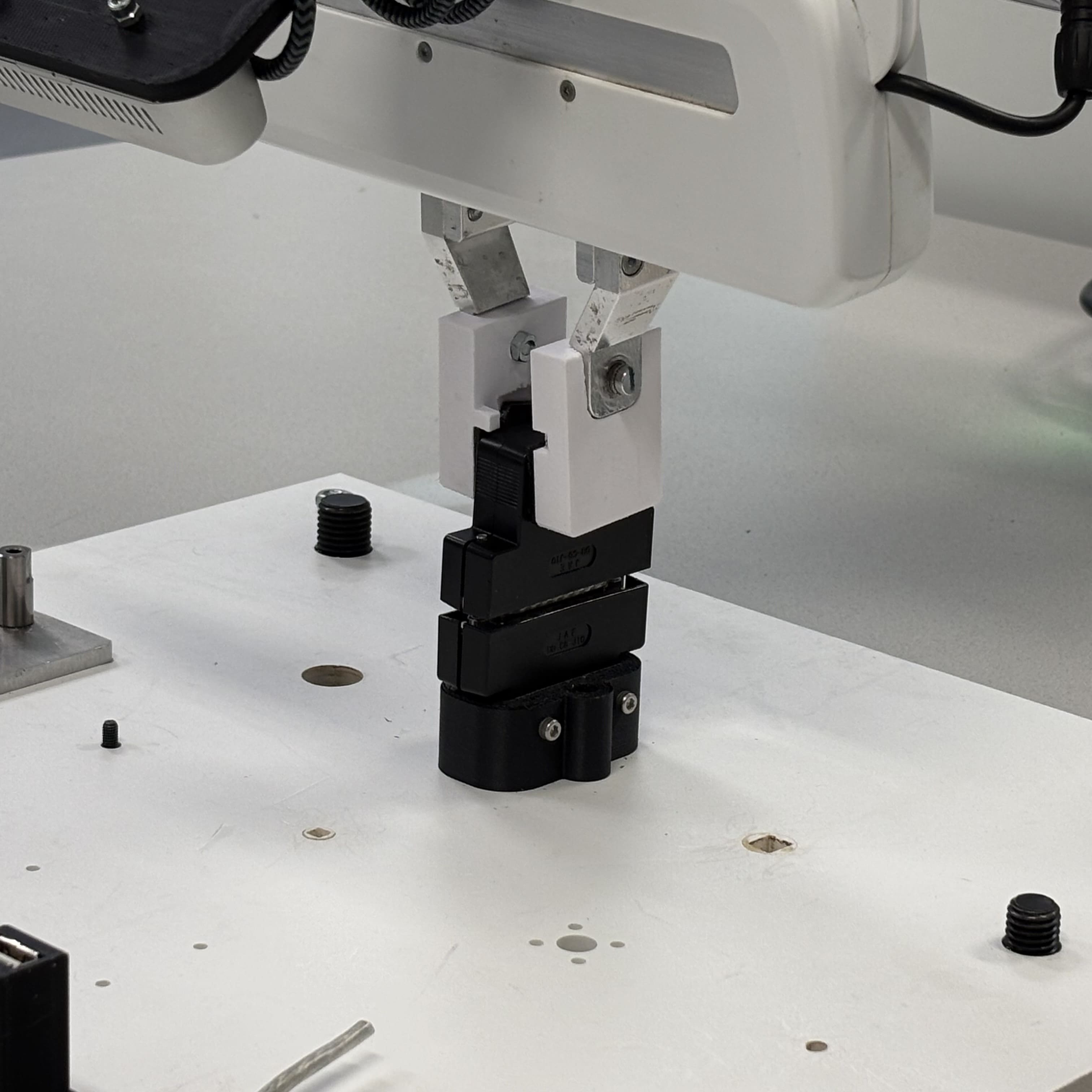}};
        \fill[white, opacity=0.1] (img.south west) rectangle (img.north east);
    \end{tikzpicture}
    \caption{Inserted}
  \end{subfigure}
  \hfill
\caption{DSUB insertion after fine-tuning with distillation, showing reactive recovery: the robot misses, retries, aligns, and successfully inserts.}
   \vspace{-0.4cm}
  \label{fig:dsub}
\end{figure}

We now focus on a clearly challenging task from Table~\ref{table_unseenbaseline}: the DSUB insertion, on which our method achieves only 50\% zero-shot success. We fine-tune our policy for 10 minutes using two approaches: tuning without distillation, and tuning with distillation from a new vision-enabled teacher pretrained on the DSUB task for 20 minutes. We also compare to the Residual RL baseline \cite{residualfromdemo}, which achieved the best results for DSUB in Table~\ref{table_unseenbaseline}, fine-tuned for 30 minutes.
Table~\ref{table_fine-tuning} shows that fine-tuning with distillation achieves a 100\% success rate. Fine-tuning without distillation, however, drops performance to zero, corroborating the findings in Section~\ref{sec ablation}: training the vision-free policy without teacher guidance is ineffective regardless of additional interaction time. Fine-tuning Residual RL for the same total duration yields only marginal improvement. Fig.~\ref{fig:dsub} depicts the fine-tuned policy executing the DSUB insertion.

\begin{table}[!tbp]
\caption{DSUB insertion success rates before and after fine-tuning. Our method fine-tunes for 10 minutes with or without distillation from a DSUB-specific teacher (20 min). Residual RL is fine-tuned for 30 minutes.}
\vspace{-0.3cm}
\label{table_fine-tuning}
\begin{center}
\begin{tabular}{l c}
\toprule
 & DSUB \\
 \midrule
Residual RL \cite{residualfromdemo} zero-shot & 6/10 \\ 
\textit{VE2VF} zero-shot & 5/10 \\ 
\addlinespace[1pt]
\arrayrulecolor{black}
\specialrule{0.3pt}{0pt}{0pt}
\arrayrulecolor{black}
\addlinespace[2pt]
Residual RL \cite{residualfromdemo} fine-tuned & 7/10 \\ 
\textit{VE2VF} fine-tuned w/o distillation & 0/10 \\ %
\textit{VE2VF} fine-tuned w/ distillation & \textbf{10/10}\\ 
\bottomrule
\end{tabular}
\vspace{-0.3cm}
\end{center}
\end{table}

\section{CONCLUSIONS}

We presented a human-in-the-loop RL framework that trains a vision-enabled teacher policy and distills it into a vision-free student policy for contact-rich manipulation. Our experiments on the NIST benchmark board demonstrate that the distilled proprioceptive policy achieves 95\% overall success across training, disturbed, and unseen tasks, outperforming all baselines. The ablation study confirms that distillation is essential, as proprioceptive policies trained from scratch fail to acquire effective insertion strategies, and that the combination of pose, twist, and wrench inputs yields the greatest robustness. For challenging tasks, fine-tuning with a task-specific teacher achieves full success within minutes of additional interaction. Future work includes extending the method to other contact-rich skills, such as screwing and non-prehensile manipulation, using language conditioning or task embeddings, and considering additional observation encoding strategies for dynamic modality prioritization.


\addtolength{\textheight}{-0cm}   





\bibliographystyle{IEEEtran}
\bibliography{bibtex}


\end{document}